\documentclass[runningheads]{llncs}

\usepackage{subfigure}
\usepackage{color}
\usepackage{algorithm}
\usepackage{algpseudocode}
\usepackage{booktabs}
\usepackage{array}
\usepackage{verbatim}
\usepackage{gensymb}

\newcommand{\etal}{\textit{et~al}\mbox{.}}

\definecolor{gray}{rgb}{0.5,0.5,0.5} 
\definecolor{green}{rgb}{0, 0.4, 0} 
\definecolor{orange}{rgb}{1, 0.5, 0} 	
\definecolor{mahogany}{rgb}{0.75, 0.25, 0.0}
\definecolor{purple}{rgb}{0.6, 0, 0.6}
\definecolor{frenchblue}{rgb}{0.0, 0.45, 0.73}

\newboolean{revising}
\setboolean{revising}{true}
\ifthenelse{\boolean{revising}}
{
	\newcommand{\ignore}[1]{}

}
{
	\newcommand{\ignore}[1]{}

}

\usepackage{times}
\usepackage{epsfig}
\usepackage{graphicx}
\usepackage{amsmath}
\usepackage{amssymb}

\usepackage{color}
\usepackage[width=122mm,left=12mm,paperwidth=146mm,height=193mm,top=12mm,paperheight=217mm]{geometry}
\usepackage[pagebackref=true,breaklinks=true,letterpaper=true,colorlinks,bookmarks=false]{hyperref}

\begin{document}

\pagestyle{headings}
\mainmatter

\title{FishNet: A Camera Localizer using Deep Recurrent Networks}
\author{Hsin-I Chen \and Sebastian Agethen \and Chiamin Wu \and Winston Hsu \and Bing-Yu Chen}
\institute{National Taiwan University}

\maketitle

\begin{abstract}

This paper proposes a robust localization system that employs deep learning for better scene representation, and enhances the accuracy of 6-DOF camera pose estimation. 
Inspired by the fact that global scene structure can be revealed by wide field-of-view, we leverage the large overlap of a fisheye camera between adjacent frames,  
and the powerful high-level feature representations of deep learning.
Our main contribution is the novel network architecture that extracts both temporal and spatial information using a Recurrent Neural Network. Specifically, we propose a novel pose regularization term combined with LSTM. This leads to smoother pose estimation, especially for large outdoor scenery.  
Promising experimental results on three benchmark datasets manifest the effectiveness of the proposed approach.

\end{abstract}

\section{Introduction}
\begin{figure}[t!]
\includegraphics[width=0.49\columnwidth]{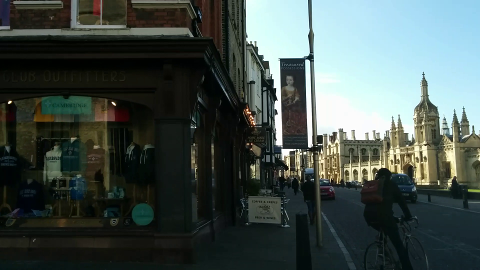}
\includegraphics[width=0.49\columnwidth]{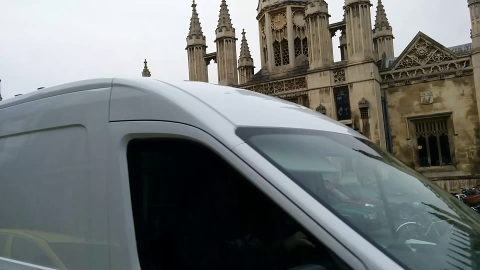}
\caption{Obstacles during localization. \textbf{Left:} Strongly varying illumination. \textbf{Right:} Short-term occlusion by passing traffic.}
\label{fig:introduction}
\end{figure}

Image-based localization, defined as the problem of estimating the position and orientation of a camera, has received substantial attention in the robotics and computer vision community. It is essential for tasks such as landmark recognition~\cite{LiSH10}, autonomous navigation~\cite{MajdikVAS14}, augmented reality~\cite{ArthPVSL15} and visual odometry~\cite{NisterNB04}. Fig.~\ref{fig:introduction} introduces some of the challenges of this problem -- small and barely visible features, occlusions and the need to be robust to perspective and illumination changes.

The main stream of work in this field has been motivated by the above challenges, the need to establish a large enough feature correspondences between a query image and the 3D scene.
Structure-based approaches~\cite{IrscharaZFB09,SattlerLK12,SattlerHRSP15,LiSH10} associate image descriptors, e.g., SIFT, with 3D scene points during \emph{Structure-from-Motion} (SfM) reconstruction.
This representative set of 3D points that cover a 3D scene from arbitrary viewpoints allows better registration of query images taken from novel viewpoints.
The above effectiveness, however, is achieved at the cost of limited expressiveness -- the use of handcrafted local features
, and the lack of scalability due to the increasing complexity of 3D scenes and the large memory footprint required for local descriptors.

Given the recent progress in deep learning, several methods proposed the use of Convolutional Neural Networks (CNNs) to learn feature representations for image localization. The central idea is to leverage transfer learning from recognition to re-localization, followed by formulating the pose estimation as a regression~\cite{kendall2015convolutional} or as a classification~\cite{Weyand16PlaNet} problem. 
CNNs have the advantage of the availability of high-level features, while simultaneously reducing the memory consumption.
Despite that, the current approaches suffer from at least two shortcomings. 
First, commonly used transfer learning models are pretrained on unrelated datasets, e.g., classification on ImageNet~\cite{ILSVRC15}, which may be suboptimal for localization. Besides, without explicitly reconstructing the 3D scene, the lack of global structure information inhibits the network from learning better spatial representations.
Second, valuable temporal information is not exploited, causing such approaches prone to failure in the presence of short-term noise.    

We aim to address the two aforementioned problems simultaneously. Our approach relies on the insight that
tracking visual landmarks over longer periods of time allows 
to discover more scene structure information.
Based upon this, the accuracy of deep pose estimation increases with the availability of sequences of measurements. 
Complementing that, a wider field-of-view allows 
larger visual overlap between subsequent images. 
This implies that deep learning can simultaneously increase robustness as the visual overlap between subsequent images is larger.  
We leverage this property and increase the field-of-view from both spatial and temporal dimension for 6-DOF camera localization. 
Specifically, we capture the wider spatial information encoded
in multiple adjacent frames using recurrent networks, inspired by~\cite{Weyand16PlaNet} and increase the robustness using fisheye camera. 

The proposed approach is built upon a Deep Convolutional Network (DCN). By employing \emph{Long Short Term Memory} (LSTM), a type of recurrent network, we can then leverage the information present in consecutive frames of a sequence. During learning and inference, high-level, pretrained GoogLeNet DCN features are used as input to the LSTM. Temporal coherency of the output pose sequence is enforced by adding a new regularization term to the commonly used regression loss. Between frames, the observer typically moves a small distance only, and further regularization can help to avoid sudden jumps due to noise, such as short-term occlusion. 

The resulting model is a scene-specific recurrent network that performs pose regression on image sequences captured by a monocular camera. 
There are two advantages of this formulation. 
First, the CNN with recurrent network is capable of exploiting temporal dependencies, therefore uncertainty is reduced. We suggest that -- due to the tendency of the fisheye model to project the scene to a wide range of spatial context -- such a regularized system with fisheye cameras works particularly well for outdoor scenes. Second, the approach is substantially simpler to formulate than methods based on 2D-to-3D models. 

In this work, we apply the presented model to the task of image-based localization. 
Finally, our approach is comprehensively evaluated on two real-world scenarios, one large outdoor and one indoor scene dataset, as well as one synthetic that includes various camera optics. We also provide comparison with a structure-based approach. Experiments shows that our method is robust and able to localize the images.

\paragraph{Contribution} 
We investigate the problem of camera localization using a deep learning approach. In summary, our work makes the following contribution:
\begin{itemize}

\item We are the first to use fisheye imagery in deep learning to increase the field-of-view for image localization. 
\item We propose a new LSTM architecture with a novel regularization term, of which fisheye lens cameras in outdoor scenes profit particularly. 
\item We provide a quantitative comparison of structure-based and learning-based localization approaches and show significant improvements over PoseNet~\cite{kendall2015convolutional}. Experiments show that increasing the camera field-of-view together with our architecture has a significant impact for image localization.
\end{itemize}

\section{Related Work}

\paragraph{Localization from Structure}
Detail information obtained from a 3D reconstruction of the scene is essential to achieving high localization accuracy~\cite{IrscharaZFB09,LiSHF12,ZeislSP15,Sattler16}. 
The common pipeline is to use the features descriptors for the 3D points computed during structure from motion~\cite{schoenberger2016sfm}, formulating the correspondences search as a descriptor matching problem~\cite{Sattler11FIL}. The $6$-DOF camera pose of a query image can be estimated from the correspondences between 2D local features in the query image and 3D points in the model using camera resectioning. 
Some notable works have brought about significant progress. Sattler~\etal~\cite{SattlerLK12} developed prioritized search strategies for efficient 2D-to-3D correspondences search. To further accelerate descriptor matching, a model compression scheme by means of quantizing the point descriptors is introduced in~\cite{SattlerHRSP15}. 
Although explicitly constructing a 3D model of the structure aids determining the camera's poses, the cost of memory storage and computations gets more expensive as the size of the 3D model grows. Our solution can leverage the strength of the 3D structure but does not need to store it explicitly.  

\paragraph{Localization from Learning}
Recent advances in camera localization use predictions from a regression forest to guide the camera pose optimization procedure. Shotton~\etal~\cite{ShottonGZICF13} employ a regression forest to infer the pose of a RGB-D camera. Valentin~\etal~\cite{Guzman-RiveraKGSSFI14} train a regression forest to predict mixtures of anisotropic 3D Gaussians. To show that 6D pose estimation can be acomplished using a single RGB image, Brachmann~\etal~\cite{BrachmannMKYGR16} further marginalize the object coordinate distributions over depth. However, these approaches require depth information during training. 
Thus, they are better suited for indoor applications.

\paragraph{Deep Convolutional Network}
Deep learning is being used for a wide array of computer vision tasks, such as image classification~\cite{KrizhevskySH12,googlenet16,lit:vgg}, object localization~\cite{SermanetEZMFL13} and detection~\cite{lit:faster_rcnn}, as well as segmentation~\cite{HariharanAGM14,GirshickDDM14}. Deep Convolutional Networks (DCN), as first being used in~\cite{LeCunBoserDenkerEtAl89}, have demonstrated impressive abilities at extracting high-level features and form one of the pillars of deep learning. At the same time, we are now able to train networks deeper than ever before with the help of GPU-based training. 

Time-variant data is a particular challenge due to its increased size. Nonetheless, it also gives us the opportunity to extract additional useful information. Recent years have seen the use of Recurrent Neural Networks (RNN) to process such data, of which Long Short Term Memory (LSTM) is particularly popular. Previous applications for such sequence learning include video classification~\cite{lit:beyondshortsnippets}, natural language generation and processing for image and video captioning~\cite{lit:imagecaptioning,lit:videocaptioning}, and future prediction~\cite{lit:futureprediction}.

In this work, we employ both techniques for the task of camera localization. We use the fact that sequences of video frames can be used by employing RNNs in order to improve localization and reduce impact of intermittent, short-term noise. 

\paragraph{Fisheye Camera} has received increasing attention with its broad applications in 3D reconstruction and visual odometry (VO). Unlike a classical pinhole camera that shows only the front view of a scene, a fisheye camera can capture omni-directional lights from the surrounding environment.
Caruso~\etal~\cite{caruso2015_omni_lsdslam} proposed a direct monocular SLAM method for wide field-of-view cameras. Im~\etal~\cite{im2016all} introduced a 3D reconstruction method for stereo-scopic panorama using spherical camera. 
Zhang~\etal~\cite{ZhangRFS16} studied the impact of different FoV on standard VO module, and show that it can benefit from large field-of-view. Inspired by their work, we also utilized fisheye-lens, characterized by large field-of-view, in order to allow our system to learn more global structure, resulting in more accurate
registration that adheres closely to the underlying scene geometry.





\section{Model Architecture}
\begin{figure}[t!]
\centering
\def\svgwidth{.66\columnwidth}
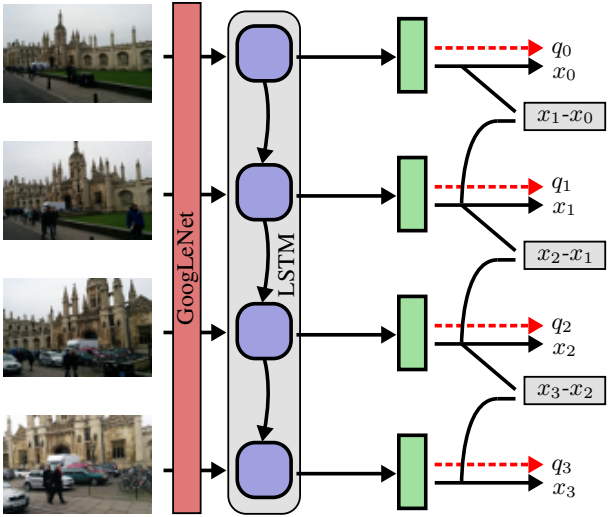
\caption{System overview. Raw Images of a sequence are fed into GoogLeNet, and high-level features are extracted. A LSTM then processes the sequence (here: $T=4$) in order. The output is used by the regressor (green) to find the position and orientation, which are learnt using an L2-loss. An additional loss (grey box) ensures that neighboring frames have similar results.}
\label{fig:architecture}
\end{figure}

We begin with a brief introduction of Long Short Term Memory (LSTM) \cite{LSTM}, and then give a formal description of our architecture in Section~\ref{sec:Model}. An overview of the proposed scheme for pose regression is shown in Figure~\ref{fig:architecture}.

\subsection{Long Short Term Memory}
Consider an input sequence $\mathbf{X} = (\mathbf{x}_0,\cdots,\mathbf{x}_{T-1})$, of length $T$, where $\mathbf{x}_t$ represents the $t$-th element. Such a sequence may for example be the RGB frames of a video clip, or features extracted from a deep convolutional stack.

Given this input, a LSTM produces a time-dependent output $\mathbf{h}_t$ by repeatedly updating its \emph{cell state} $\mathbf{c}_t$. The cell state is manipulated with the help of two \emph{control gates}, the \textit{input gate} $\mathbf{i}_t$, and the \textit{forget gate} $\mathbf{f}_t$, while an \textit{output} gate $\mathbf{o}_t$ controls the output hidden state $\mathbf{h}_t$. The LSTM equations are:
\begin{flalign}
  &\mathbf{i}_t  =  \sigma\left( \mathbf{W}_{xi}\mathbf{x}_t + \mathbf{W}_{hi}\mathbf{h}_{t-1} + \mathbf{W}_{ci} \circ \mathbf{c}_{t-1} + b_i \right)& \\
  &\mathbf{f}_t  =  \sigma\left( \mathbf{W}_{xf}\mathbf{x}_t + \mathbf{W}_{hf}\mathbf{h}_{t-1} + \mathbf{W}_{cf} \circ \mathbf{c}_{t-1} + b_f \right)& \\
  &\mathbf{o}_t  =  \sigma\left( \mathbf{W}_{xo}\mathbf{x}_t + \mathbf{W}_{ho}\mathbf{h}_{t-1} + \mathbf{W}_{co} \circ \mathbf{c}_{t-1} + b_o \right)& \\
  &\mathbf{c}_t  =  \mathbf{f}_t\circ\mathbf{c}_{t-1}+\mathbf{i}_t \circ \tanh\left(\mathbf{W}_{xc}\mathbf{x}_t + \mathbf{W}_{hc}\mathbf{h}_{t-1} + b_c\right)& \\
  &\mathbf{h}_t  =  \mathbf{o}_t \circ \tanh\left(\mathbf{c}_t\right)&
\end{flalign}
where $\sigma$ is the logistic sigmoid function, and $\mathbf{W}, b$ are the parameters of the LSTM model. We term the operations $\mathbf{W}_{x*} \cdot \mathbf{x}_t$ the \emph{input-to-hidden} transition, and the operations $\mathbf{W}_{h*} \cdot \mathbf{h}_{t-1}$ the \emph{hidden-to-hidden} transition.
We also remark here that some literature and implementations may ignore the hadamard terms $\mathbf{W}_{c*} \circ \mathbf{c}_{t-1}$. 


\subsection{Camera Pose Regression with LSTM}
\label{sec:Model}
Given an input sequence $\mathbf{X}$, for each frame $\mathbf{x}_t$, our network outputs a pose vector $\mathbf{P}_t$ , which can be seperated into a camera position $\mathbf{p}\in\mathbb{R}^3$ and an orientation represented by a quaternion $\mathbf{q}\in \mathbb{R}^4$:
\begin{equation}
\mathbf{P}=[\mathbf{p},\mathbf{q}]
\end{equation}

We adopt GoogLeNet~\cite{googlenet16} to process the input images, and extract $1024$-dimensional features at \texttt{pool5}. These extracted feature map serve as input to a LSTM unit. At each timestep $t$, we first apply the dropout~\cite{lit:dropout} technique on the hidden state $\mathbf{h}_t$ of LSTM layer. It then serves as input to a $7$-dimensional pose regressor, which is implemented as a fully-connected layer. The pose regressor then outputs the desired pose $\mathbf{P}$.

\subsection{Network Loss Function}

Given the training images and their ground truths, the euclidean loss is described as followed~\cite{kendall2015convolutional}:
\begin{equation}
L = \left\|\hat{\mathbf{x}}-\mathbf{x}\right\|_2 + \beta \left\|\hat{\mathbf{q}}-\frac{\mathbf{q}}{\left\| \mathbf{q}\right\|}\right\|_2,
\end{equation}
where $\hat{\mathbf{x}}$ and $\hat{\mathbf{q}}$ represent the ground truth label for each image, and $\beta$ is a scale factor that balances the loss between the location error $\left\|\hat{\mathbf{x}}-\mathbf{x}\right\|_2 $ and orientation error $\left\|\hat{\mathbf{q}}-\frac{\mathbf{q}}{\left\| \mathbf{q}\right\|}\right\|_2$. 
In addition, an L2-regularized weight decay -- scaled by $\gamma$ -- is added, as it helped the network generalize better. Following the notation of \cite{kendall2015convolutional}, we omit the normalization term $\|\mathbf{q}\|$ in the following.

\paragraph{Sequence Learning: Regularization}
In our work, we consider sequences of length $T$ as input.
A sequence is a collection of consecutive frames, and as such, it should be expected that the difference between frame $t$ and $t+1$ is small. To enforce this, we can add an additional temporal regulization term, weighted by $\delta$, to our loss function. The total loss function for image $t$ is:

\begin{multline}
  L_t = \left\|\hat{\mathbf{x}_t} - \mathbf{x}_t\right\|_2 + \beta \left\|\hat{\mathbf{q}}_t - \mathbf{q}_t\right\|_2 \\ + \gamma \left\| \theta \right\|_2 + \delta \left\|\mathbf{x}_t - \mathbf{x}_{t-1}\right\|_2, \quad t \geq 1,
\label{eq:loss}
\end{multline}

where $\theta$ are the network's parameter. We define the temporal regulization term $\delta\|\mathbf{x}_t - \mathbf{x}_{t-1}\|$ to be zero for $t=0$. 

The total loss for a sequence is then the summation over time:
\begin{equation}
L_{\text{total}} = \sum\limits_{t=0}^{T-1} L_t
\end{equation}

For a fisheye lens, which is characterized by a large field-of-view, compared to the classical perspective camera, the image projections are smaller under the same image resolution. Thus it is much more suitable for our regualization term since smaller distance and frame rate is required.

\section{Experimental Results}
In this section, we illustrate our experiments on the aforementioned architecture. We begin with implementation details, compare our system with the established baseline in~\cite{kendall2015convolutional} and structure-based approach in~\cite{Sattler11FIL}, and conclude with an evaluation of Fisheye and Perspective cameras on a synthetic dataset.

\paragraph{Datasets}
The performance of the proposed method and various related techniques are evaluated on three publicly available datasets, including the \emph{Cambridge Landmark} dataset~\cite{kendall2015convolutional}, the \emph{7-Scenes} dataset~\cite{ShottonGZICF13}, and the \emph{Multi-FoV} synthetic dataset~\cite{ZhangRFS16}. These sequences exhibit depth variations, contain dynamic moving objects and different spatial content, and thus are very challenging for image-based localization.

The Cambridge Landmark dataset contains outdoor sequences. The appearance of large spatial content, partial occlusions, and urban clutter make localization over this dataset quite challenging. However, it provides a good test-bed to manifest the importance of temporal information (recurrent network), as localization approaches that rely on a single images are not reliable enough. 
To test indoor scenes, we use the publically available \textit{7-Scenes} dataset, which contains large variation in camera height. 

\paragraph{Implementation details}
Our proposed architecture can be seperated into a CNN feature extractor based on the Inception architecture (GoogLeNet)~\cite{googlenet16}, and the recurrent network with regressor. The GoogLeNet model is initialized with weights pretrained on the ImageNet 1K image classification dataset. The remainder of the network is intialized with random values, where we use Xavier initialization for the recurrent net, and a Gaussian initialization for the regressor. In case of Cambridge, the outdoor dataset, we choose $\sigma=0.5$ to initialize the position regressor, as it needs to regress large coordinates. For 7-Scenes, we choose $\sigma=0.1$, and the orientation regressor is always initialized with $\sigma=0.01$. During training, we minimize the Euclidean loss of Eq.~\ref{eq:loss} using the Adam~\cite{KingmaB15} optimizer. A pixelwise mean is subtracted for each image. All networks take image crops of size $224\times224$ as input. During training, the crops are chosen at random positions in the image, while at test time a crop around the center point is being used. We set the hyperparameters as follows: batch size $20$, i.e., processing $T\times20$ images in parallel, dropout probablity of $0.5$, weight decay coefficient $\gamma = 0.0002$, and temporal regularization coefficient $\delta = 0.0002$. To find the trade-off constant $\beta$, which regulates position vs. orientation learning, we follow \cite{kendall2015convolutional} and set it such that the magnitude of both loss terms is about equal. If not otherwise mentioned, we use sequences of $T=3$ frames. All experiments are performed on an NVIDIA Tesla K80 GPU.

\begin{table*}[t]
\centering
\caption{Localization result (median error) on Cambridge Landmark dataset. Note that \emph{LSTM (Reg.)} refers to our proposed network with regularization.}
\label{tbl:cambridge}
\begin{tabular}{|p{2.82cm}|p{1.9 cm}|p{2.0cm}|p{2.0cm}|l|l|}
\hline
Scene & 2D-3D Matching~\cite{SattlerLK12} & PoseNet~\cite{kendall2015convolutional} & Bayesian PoseNet~\cite{kendall2015modelling} & LSTM & LSTM (Reg.) \\
\hline 
\hline
\textsc{Kings's College}  &$1.42m, 6.20\degree$    &  $1.92m,5.92\degree$ & $1.74m,4.06\degree$ & $1.14m, 3.68\degree$ & $0.98m,3.93\degree$\\
\textsc{ShopFacade}       &$0.38m, 5.63\degree$  &  $1.46m,8.08\degree$ & $1.25m,7.54\degree$ & $1.08m, 4.91\degree$ & $1.14m,5.62\degree$\\
\textsc{OldHospital}      & --      &  $2.31m,5.38\degree$ & $2.57m,5.14\degree$ & $2.29m, 4.08\degree$ & $2.27m,3.87\degree$ \\ 
\textsc{St Mary's Church} &  $0.74m, 2.19\degree$  &  $2.65m,8.48\degree$ & $2.11m,8.38\degree$ & $1.58m, 6.65\degree$ & $1.63m,6.07\degree$\\
\hline
\textsc{Average}          & $0.84m,4.67\degree$ &$2.08m, 6.38\degree$ & $1.92, 6.28\degree$ & $1.52m, 4.83\degree$ &$1.50m, 4.87\degree$ \\
\hline
\end{tabular}
\end{table*}

\begin{table*}[t]
\centering
\caption{Localization result on 7-Scenes dataset}
\label{tbl:seven_scenes}
\begin{tabular}{|p{2.cm}|p{2.cm}|p{2.0cm}|p{2.0cm}|l|l|}
\hline
Scene & 2D-3D Matching~\cite{SattlerLK12} & PoseNet~\cite{kendall2015convolutional} & Bayesian PoseNet~\cite{kendall2015modelling} & LSTM & LSTM (Reg.)\\
\hline 
\hline
\textsc{Chess}          & $0.11m,6.20\degree$  &  $0.32m,8.12\degree$ & $0.37m,7.24\degree$ & $0.20m,6.17\degree$   & $0.20m, 6.36\degree$ \\
\textsc{Fire}           & $0.34m,3.92\degree$  &  $0.47m,14.4\degree$ & $0.43m,13.7\degree$ & $0.43m,14.65\degree$  & $0.35m, 15.00\degree$ \\
\textsc{Heads}          & $0.44m,4.54\degree$  &  $0.29m,12.0\degree$ & $0.31m,12.0\degree$ & $0.24m,14.36\degree$  & $0.24m, 13.65\degree$\\
\textsc{Office}         & $0.23m,2.28\degree$  &  $0.48m,7.68\degree$ & $0.48m,8.04\degree$ & $0.41m,9.08\degree$   & $0.41m, 9.2\degree$ \\
\textsc{Pumpkin}        & $0.48m,2.06\degree$  &  $0.47m,8.42\degree$ & $0.61m,7.08\degree$ & $0.39m,6.76\degree$   & $0.36m, 6.24\degree$ \\
\textsc{Red Kitchen}    & --                   &  $0.59m,8.64\degree$ & $0.58m,7.54\degree$ & $0.43m,8.74\degree$   & $0.41m, 8.71\degree$\\
\textsc{Stairs}         & --                   &  $0.47m,13.8\degree$ & $0.48m,13.1\degree$ & $0.38m,11.72\degree$& $0.43m,11.39\degree$\\
\hline
\textsc{Average} & $0.32m,3.8\degree$ &  $0.44m, 10.4\degree$  &  $0.47m, 9.81\degree$ & $0.35m,10.21\degree$ &  $0.34m, 10.07\degree$\\
\hline
\end{tabular}
\end{table*}

\subsection{Comparison with deep learning approach}
We first run our proposed LSTM architecture and compare it with PoseNet, which does not employ sequence learning. Our results can be found in Table~\ref{tbl:cambridge} and~\ref{tbl:seven_scenes}.

We are able to improve localization accuracy and pose estimation for all datasets except \textsc{Street}, on which the network was not able to generalize, i.e., \textsc{Street} test set results remained near those of random initialization. It shows an improvement over PoseNet of up to $49\%$ in \emph{Cambridge} and $37\%$ in \emph{7-Scene}.
Results on the Cambridge dataset, which is an outdoor dataset, tend to show greater improvement than for the 7-Scene dataset, which is indoors.

Adding temporal regulization showed some promising results, in particular on \textsc{KingsCollege}, where location accuracy was improved by an additional 0.16m (14\% relative to LSTM) . We were not able to improve results using this technique on all datasets however. One issue is the hyperparameter $\delta$, which requires adjustment according to the distance in position of two consecutive frames: A sequence with large differences in position between frames requires a very small values of $\delta$, while sequences with minimal frame-by-frame differences may allow larger values.

\paragraph{Memory Footprint \& Performance} Using batches of 60 sequences, the average processing time per frame was 16.99 ms during training, and 9.5 ms during inference. The total number of parameters is approximately 13.8 million, of which 5.79 million are GoogleNet convolutional parameters.

\begin{figure*}[t]
  \includegraphics[width=0.5\columnwidth]{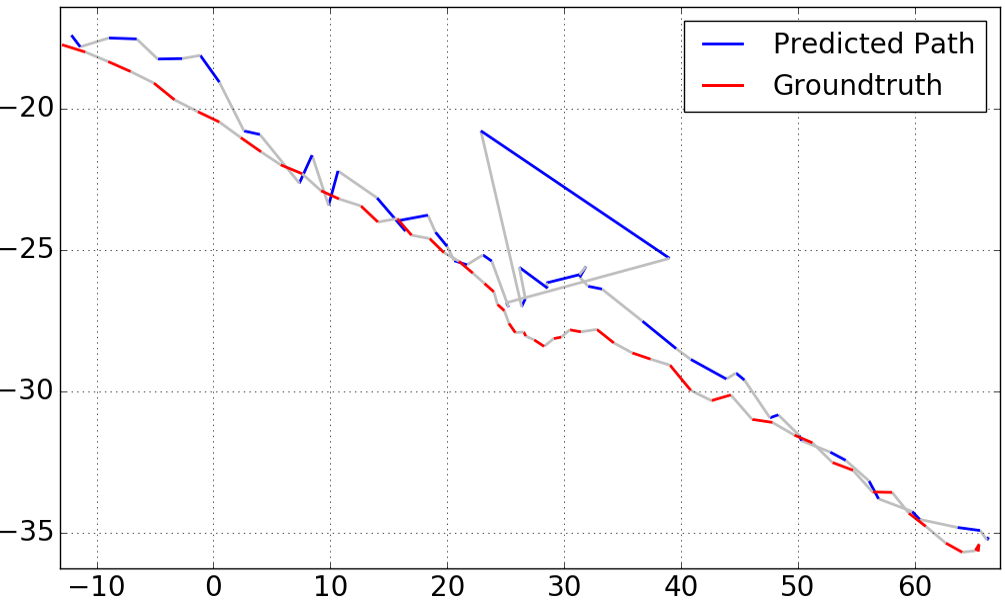}\hspace{0.05in}
  \includegraphics[width=0.5\columnwidth]{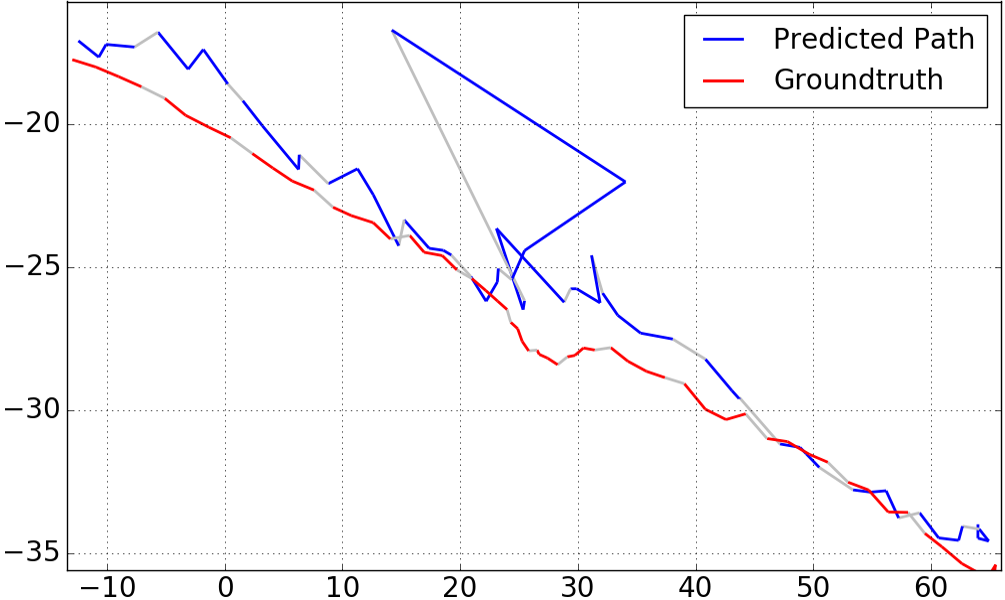}\hspace{0.05in}
  \includegraphics[width=0.5\columnwidth]{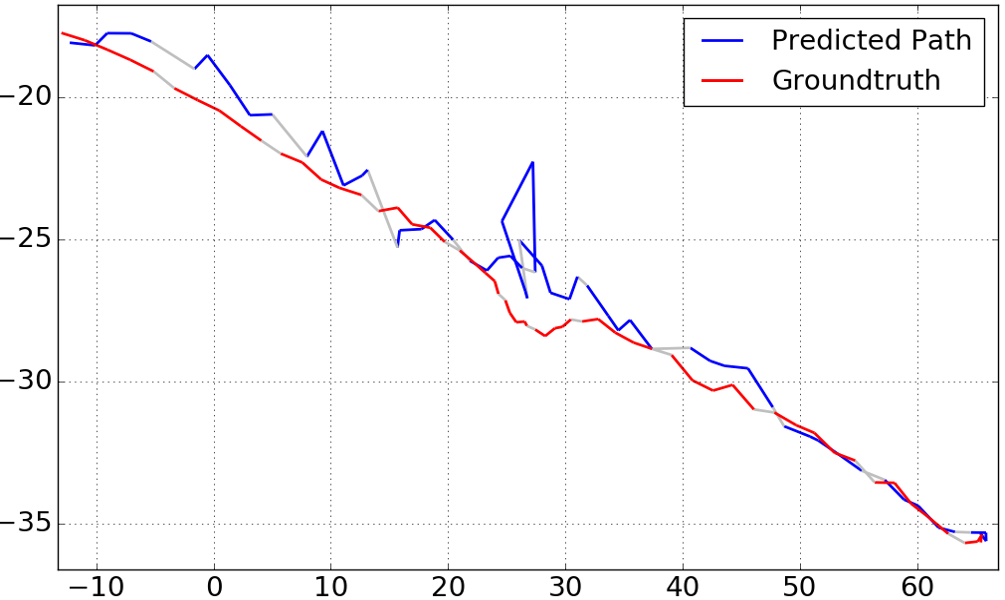}\hspace{0.05in}
  \includegraphics[width=0.5\columnwidth]{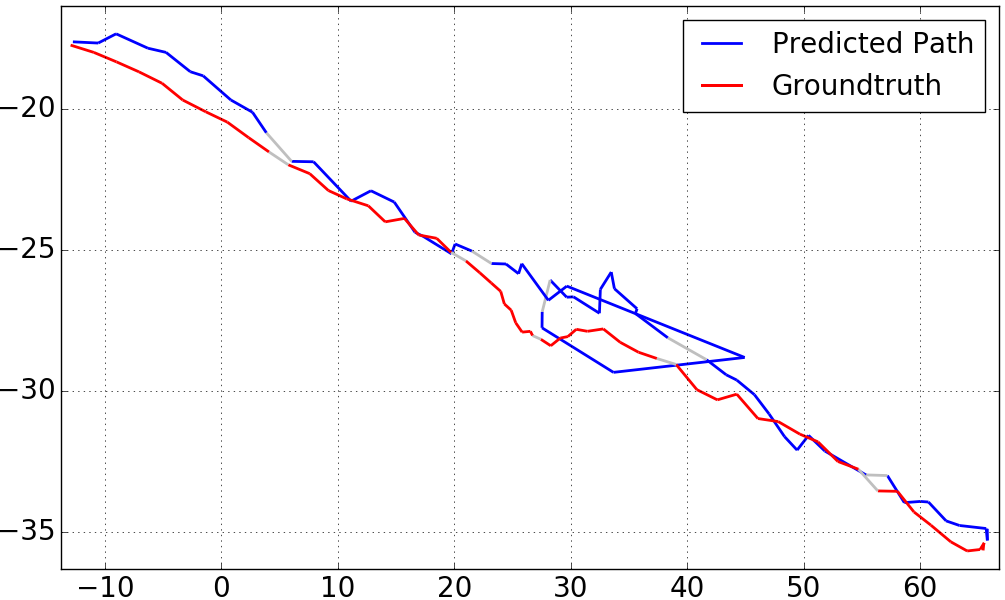}
  \caption{From left to right: Test results on sequence 2 of \textsc{KingsCollege} for $T=2,4,5,10$. Whereas a system with small values of $T$ is susceptible to short-term noise, choosing too large values of $T$ degrades the overall performance of the system.}
  \label{fig:paths_other_T}
\end{figure*}

\subsection{Comparison with 2D-to-3D descriptor matching}
In this section, we compare our apporoach with the structure-based approach~\cite{Sattler11FIL} for image localization. 
In~\cite{Sattler11FIL}, a 2D-to-3D descriptor matching is used to estimate the camera poses w.r.t a SfM model, where each 3D point is represented by a SIFT descriptor of training images obtained from the reconstruction. As the \emph{Cambridge Landmark} and \emph{7-Scenes} datasets do not contain both feature descriptors and reconstructed 3D model, we extract SIFT descriptors~\cite{Lowe04distinctiveimage} and reconstruct the 3D scene and camera path using Visual-SFM~\cite{Wu13,WuACS11}. We then register the generated camera poses to the ground truth poses in case of the Cambridge Landmark dataset~\cite{kendall2015convolutional}. 

We train a visual vocabulary containing $10$K words on the Cambridge Landmark dataset and a vocabulary of $1$K words on the 7-Scenes dataset. 
We follow~\cite{Sattler11FIL} and accept a query image as registered if the best poses estimated by RANSAC~\cite{AICPub836} from the established 2D-to-3D correspondences have at least $12$ inliers. The camera pose is estimated using the standard $6$-point DLT algorithm~\cite{Hartley2004}. 
For the implementation, we use the released code from the author's of~\cite{Sattler11FIL} website\footnote{https://www.graphics.rwth-aachen.de/software/image-localization}.

Our results are reported in Table~\ref{tbl:cambridge} as well as Table~\ref{tbl:seven_scenes}. We found that 2D-to-3D matching consistently produces smaller error on the outdoor dataset. The comparison results validate that the reconstructed 3D point clouds provide global scene structure information required for camera registration.
Note that we do not report the localization result on OldHospital because of a corrupted 3D model. 
 
\vspace{-1em}
\subsection{Comparison of Fisheye and Perspective camera}

\begin{figure}
  \centering
  \includegraphics[width=.49\columnwidth]{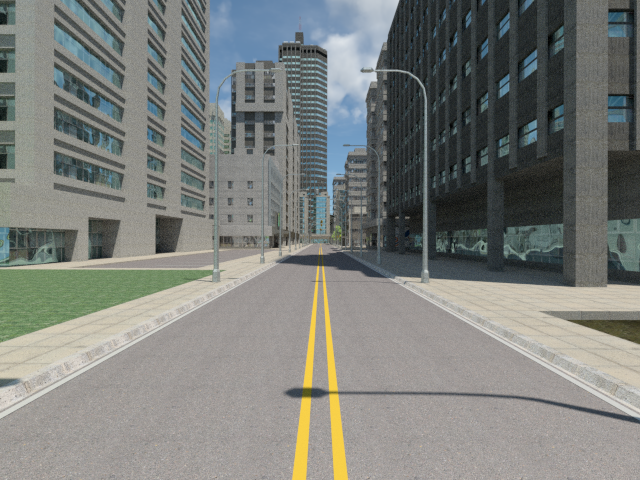}
  \includegraphics[width=.49\columnwidth]{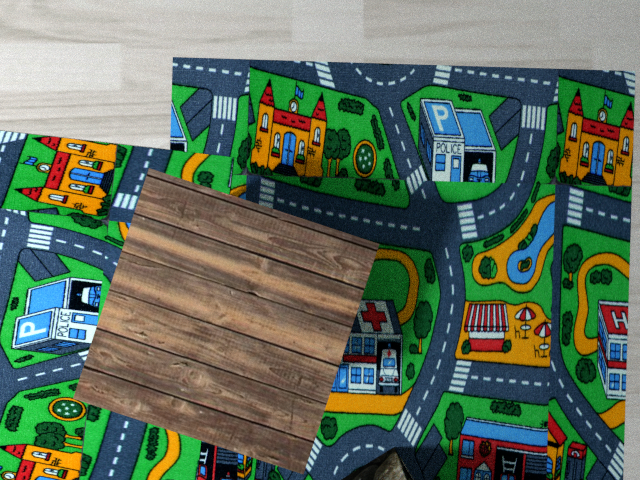}
  \includegraphics[width=.49\columnwidth]{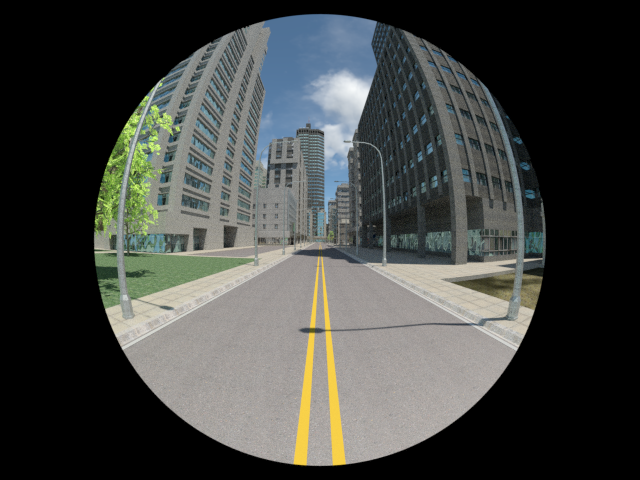}
  \includegraphics[width=.49\columnwidth]{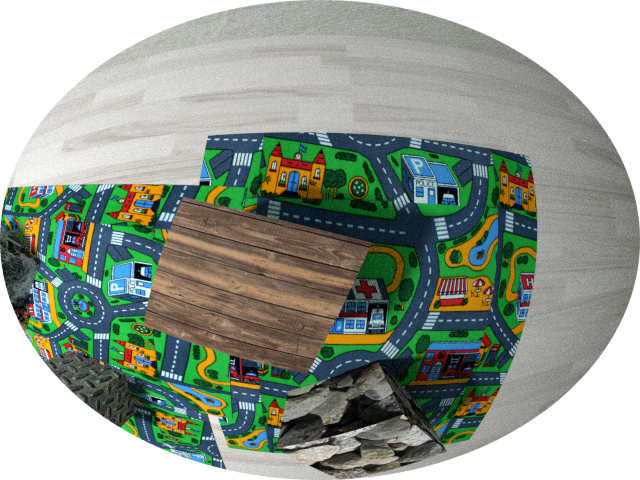}
  \includegraphics[width=.49\columnwidth]{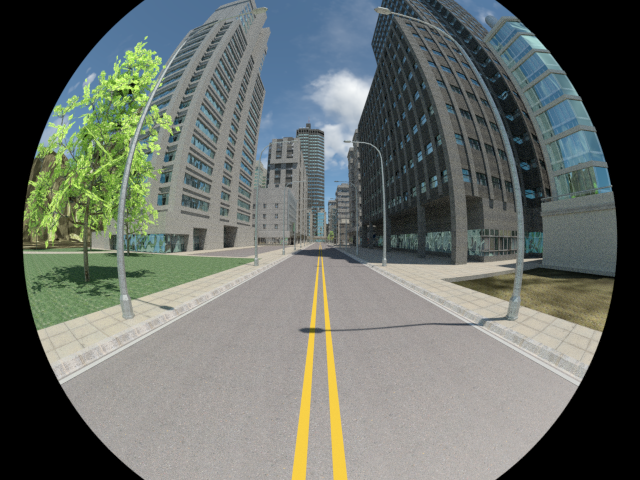}
  \includegraphics[width=.49\columnwidth]{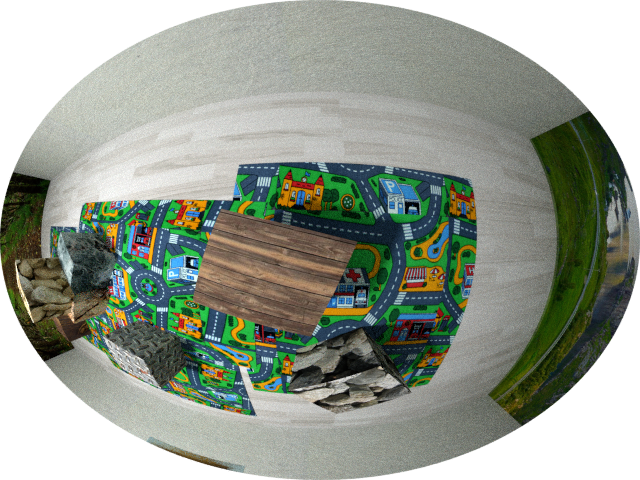}
  
  \caption{Images of the synthetic \emph{Urban Canyon} (left) and \emph{Indoor} (right) datasets. From top to bottom: The three different FoV cameras (Perspective, Fisheye $130\degree$, Fisheye $180\degree$).}
  \label{fig:synthetic}
\end{figure}

In the following, we evaluate our method and contrast its behavior for a perspective and a fisheye camera.

\subsubsection{Multi-FoV Dataset}
To allow a fair comparison, we make use of two synthetic sequences, \emph{Urban Canyon} and \emph{Indoor}~\cite{ZhangRFS16}. The use of synthetic sequences enables us to simulate different camera optics without having any variations in path or pose. That is, for both evaluated cameras, the image data only differs from use of a different camera model. We create three instances of each sequence with varying degree of field of view: Perspective $(90\degree)$, and Fisheye ($130\degree, \text{ and } 180\degree$).

On the \emph{urban canyon} dataset, we fix the camera height at $1.6m$ to simulate a pedestrian filming and walking throuh streets. For the \emph{Indoor} dataset, we fix the camera height at $1.1m$ to simulate a drone flying around. The image dimensions are kept at the VGA size ($640\times480$) which is a typical size of the recordings of dashcams, robotic sensors and drone cameras. The fisheye images suffer from severe distortion in the four corners of the image due to the nature of the projection, and we fill these regions such that -- after subtraction of mean -- the values equal $0$. An example of our perpspective and fisheye images are shown in Figure~\ref{fig:synthetic}.

\subsubsection{Fisheye analysis}

\begin{figure}[t]
  \centering
  \includegraphics[width=1.0\columnwidth,height=2in]{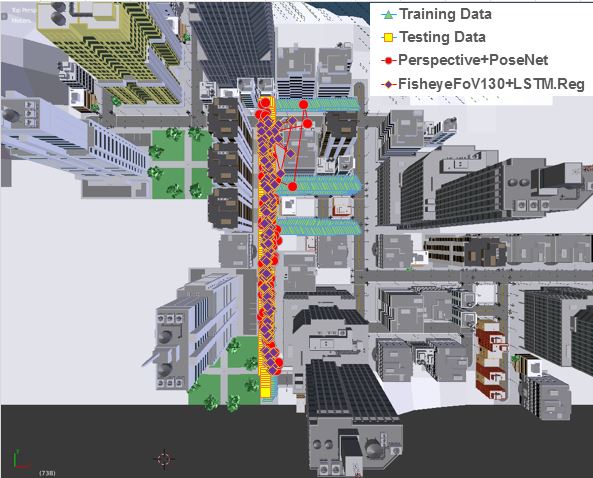}
  \includegraphics[width=1.0\columnwidth,height=2in]{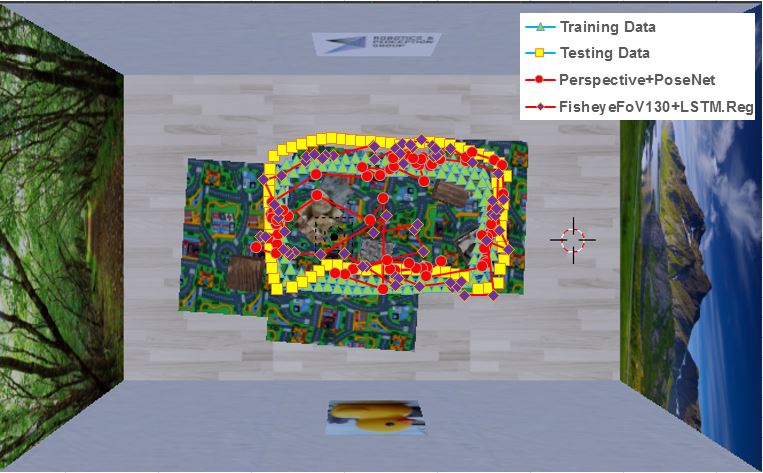}
  \caption{Visualization of camera paths of \emph{urban canyon} (left) and \emph{indoor} (right) sequences. It shows training frames (light blue), ground truth of testing frames (yellow), the predicted camera paths for perspective-90 + PoseNet (red) and the Fisheye-130 + LSTM-Reg (purple). }
  \label{fig:synthetic_path}
\end{figure}

We present our results for the two sequences in Table \ref{tbl:fisheye_city} and Table \ref{tbl:fisheye_indoor}. The addition of LSTM, that is, the use of sequence learning, improves both location and orientation accuracy considerably. A improvement up to $25\%$ on FoV-130 Urban Canyon and $22\%$ on FoV-130 Indoor.
 However, in this section we explicitly focus on the difference in learning on Perspective camera images and Fisheye camera images. Here, we can see a trend that learning on the Fisheye camera images profited from our regularization scheme, especially on the outdoor \emph{Urban Canyon} dataset. We hypothesize that the additional overlap between consecutive images due to the larger field of view allows the regularized LSTM to access its full potential.
Considering that the performance of pure LSTM slightly degraded, while the regularized system improved, it can be said that the influence by the distortion can be controlled by the regularization term. In addition, we note that, although the underlying GoogLeNet model has only been trained on perspective imagery, it was possible to adapt to the different optics during the learning process. 

A visualization of predicted paths for the two sequences is shown in Figure~\ref{fig:synthetic_path}. We can observe from the figure that Fisheye with $130\degree$ with our LSTM + Reg. gives significant improvement over the original PoseNet using perspective camera.

\begin{table}[t]
\centering
\caption{Localization result on Synthetic Urban Canyon dataset}
\label{tbl:fisheye_city}
\begin{tabular}{|l|l|c|p{1.75cm}|}
\hline
Scene & PoseNet~\cite{kendall2015convolutional} & LSTM & LSTM (Reg.) \\
\hline 
\hline
\textsc{Perspective}      &  $1.26m,6.85\degree$ & $1.02m,2.87\degree$ & $1.10m,3.72\degree$ \\
\textsc{Fisheye-130}  &  $1.13m,3.06\degree$ & $1.07m, 2.47\degree$ & $0.84m,2.48\degree$\\
\textsc{Fisheye-180}  &  $1.19m,5.63\degree$ & $1.05m,2.51\degree$ & $0.94m, 2.73\degree$\\ 
\hline
\textsc{Average}          &  $1.19m,5.18\degree$ & $1.04m,2.61\degree$ & $0.96m,2.97\degree$\\
\hline
\end{tabular}
\label{tab:dataset}
\end{table}

\begin{table}[t]
\centering
\caption{Localization result on Synthetic Indoor dataset}
\label{tbl:fisheye_indoor}
\begin{tabular}{|l|l|c|p{1.75cm}|}
\hline
Scene & PoseNet~\cite{kendall2015convolutional} & LSTM & LSTM(Reg.) \\
\hline 
\hline
\textsc{Perspective}     & $0.31m,3.75\degree$ & $0.27m,2.46\degree$      & $0.29m,3.41\degree$\\
\textsc{Fisheye-130} & $0.32m,5.46\degree$ & $0.26m,3.25\degree$      & $0.25m,3.34\degree$\\
\textsc{Fisheye-180} & $0.29m,5.38\degree$ & $0.28m,3.12\degree$      & $0.28m,3.10\degree$\\ 
\hline
\hline
\textsc{Average}        &  $0.30m,4.86\degree$ & $0.27m,2.94\degree$      & $0.27m,3.28\degree$\\
\hline
\end{tabular}
\label{tab:dataset}
\end{table}

\section{Qualitative Analysis}

\begin{figure}[t]
  \centering
  \includegraphics[width=\columnwidth]{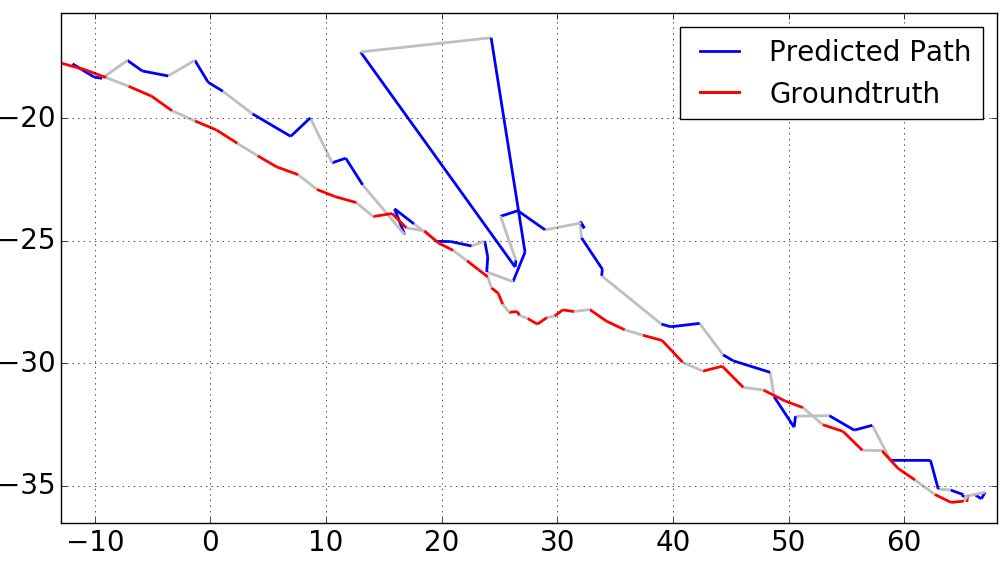}
  \caption{X- and Y-coordinates of groundtruth (red) and predicted path (blue, $T=3$) in sequence 2 of the \textsc{KingsCollege} dataset. Grey lines indicate the beginning of a new sequence. The path originates in the bottom right corner. Clearly visible is a large prediction error in the center.}
  \label{fig:result_path}
\end{figure}

\begin{figure}
  \centering
  \includegraphics[width=.75\columnwidth]{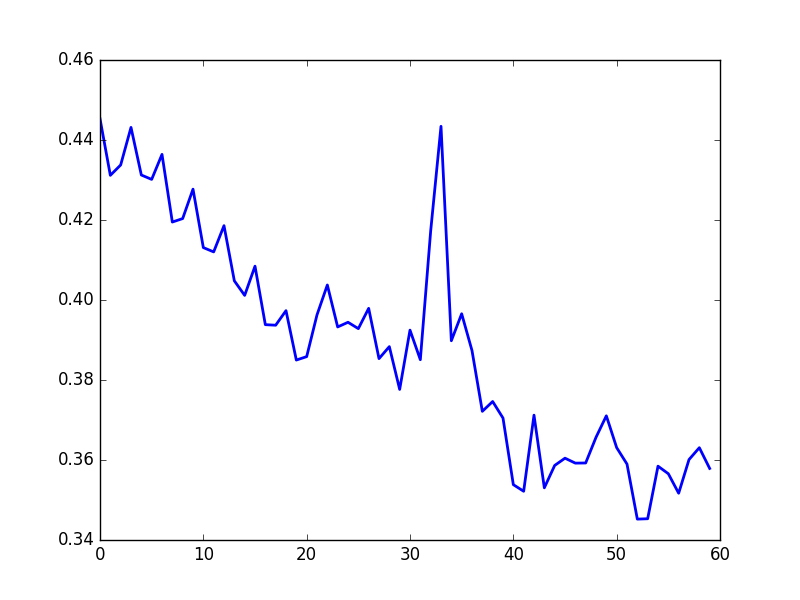}
  \caption{Median LSTM input gate activations over time (x-axis): Passing car causes a peak, thereby admitting more data to the internal state.}
  \label{fig:inputgate}
\end{figure}

\begin{figure}[t!]
  \includegraphics[width=0.49\columnwidth]{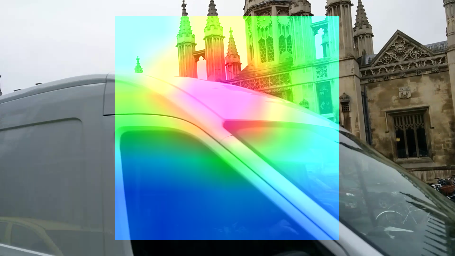}
  \includegraphics[width=0.49\columnwidth]{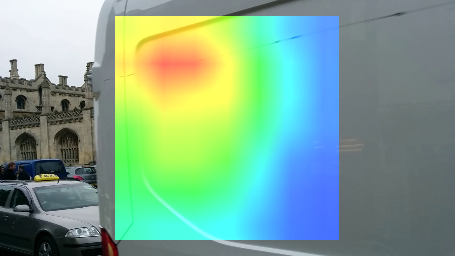}
  \includegraphics[width=0.49\columnwidth]{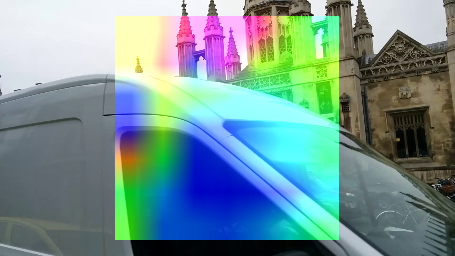}
  \includegraphics[width=0.49\columnwidth]{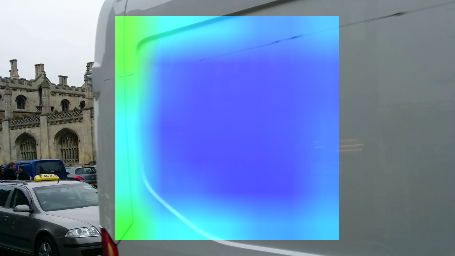}
  \caption{\textbf{Top row:} Feature activations of \emph{pretrained} GoogLeNet layer \texttt{Pool5} for center crops of frames 33, 34 of sequence 2 in the \textsc{KingsCollege} dataset. \textbf{Bottom row:} Same visualization for a network trained from scratch, here \texttt{Pool5} of AlexNet. Red areas indicate large activations, while blue indicate small activations. Notice the pretrained net recognizing something in a (from a human perspective) featureless region.}
  \label{fig:overlay}
\end{figure}

Figure \ref{fig:result_path} visualizes our results on sequence 2 of the \textsc{KingsCollege} dataset. The red segmented line represents the groundtruth, whereas the blue segments represent the result of our approach for $T=3$. A large prediction error is immediately visible in the center region, caused by frames 33 and 34 of the 60 frames sequence. Visual inspection of these frames (see Figure~\ref{fig:overlay}) reveals that the scenery is occluded by a passing car. As can be seen in Figure \ref{fig:inputgate}, the input gate of the LSTM produces an unusual high median activation at this point, and is therefore ``admitting'' more data than usual. The center crop for frame 34 exclusively shows the white car, therefore the optimal reaction of the LSTM should have been to not admit any new data.

To investigate why, we visualized the GoogLeNet \texttt{pool5} feature activation as an overlay on the RGB frames, see Figure \ref{fig:overlay}. We used the mean activations of the 1024-channel GoogLeNet output, and then repeated the same experiment with a network trained from scratch. The overlay clearly shows strong (red) activations in the top left corner.

Given the difference between a pretrained and a from-scratch network, our hypothesis is that the GoogLeNet model has been conditioned on a certain concept (for example, car-related) during pretraining on the ImageNet dataset, which now causes ``confusion'' to the LSTM unit, while the from-scratch network did not learn the concept, given that cars are an infrequent occurence.

Several mechanisms may solve this issue: Under ideal circumstances, an attention map could be used to filter out irrelevant regions, for example using the mean activations of the from-scratch network. Notice however that the features of the from-scratch networks are of lower quality. An alternative specifically for short-term noise such as passing cars could be a temporal pooling approach.

We also plot the result of sequence 2 (as seen in Figure~\ref{fig:result_path}) for other selected values of $T$. These results can be found in Figure~\ref{fig:paths_other_T}. One may expect that larger $T$ will improve the catastrophic outlier on the path, which is true as can be seen for $T=5$. We note however, that for larger values of $T$, the overall performance of the system degraded. For $T=10$, we measured a distance error of $1.2m$ and orientation error of $4.13\degree$, well above the our results for $T=3$.
\section{Conclusion}

In this paper, we presented a novel network architecture that performs camera pose regression by aggregating structure correlation from monocular image sequences. 
The proposed method leverages temporal information from adjacent frames as well as the large field-of-view revealed by fisheye. Our recurrent networks is able to deliever full $6$-DOF camera poses with high accuracy.
We apply our model to the outdoor Cambridge dataset, the indoor 7-Scenes dataset as well as a synthesic dataset, on which we evaluate fisheye versus perspective images. 
Experiments show that the proposed recurrent network architecture is able to effectively localize images compared to the previous approach.  
In our future work, we plan to investigate spatial attention masks and other mechanisms to supress short term noise.

\newpage
{\small
\bibliographystyle{ieee}
\bibliography{eccv18_deeploc}
}

\end{document}